\pdfoutput=1

\documentclass[11pt]{article}

\usepackage[]{ACL2023}

\usepackage{times}
\usepackage{latexsym}

\usepackage[T1]{fontenc}

\usepackage[utf8]{inputenc}

\usepackage{microtype}

\usepackage{inconsolata}

\usepackage{graphicx}
\graphicspath{{graphs/}}
\usepackage{amsmath}

\usepackage{physics}
\usepackage{subcaption}
\newcommand{\slack}[1]{}
\newcommand{\cready}[1]{}
\newcommand{\anonm}[1]{}
\title{ColD Fusion: Collaborative Descent for Distributed Multitask Finetuning}

\author{Shachar Don-Yehiya\\ IBM Research\\ Hebrew University of Jerusalem \\ {\small \href{mailto:shachar.don-yehiya@ibm.com}{\tt shachar.don-yehiya@ibm.com}}  
\And Elad Venezian \\ IBM Research \\ {\small \href{mailto:eladv@il.ibm.com}{\tt eladv@il.ibm.com}}\\  
\And Colin Raffel \\ UNC Chapel Hill \\ {\small \href{mailto:craffel@gmail.com}{\tt craffel@gmail.com}} \\
\AND Noam Slonim \\ IBM Research \\ {\small \href{mailto:noams@il.ibm.com}{\tt noams@il.ibm.com}}
\And Yoav Katz \\ IBM Research \\ {\small \href{mailto:katz@il.ibm.com}{\tt katz@il.ibm.com}} 
\And Leshem Choshen \\ IBM Research \\ {\small \href{mailto:leshem.choshen@il.ibm.com}{\tt leshem.choshen@il.ibm.com}}
}

\begin{document}
\maketitle
\begin{abstract}
We propose a new paradigm to continually evolve pretrained models, denoted ColD Fusion. It provides the benefits of multitask learning but leverages distributed computation with limited communication and eliminates the need for shared data. Consequentially, ColD Fusion can give rise to a synergistic loop, where finetuned models can be recycled to continually improve the pretrained model they are based upon. We show that ColD Fusion yields comparable benefits to multitask training by producing a model that (a) attains strong performance on all of the datasets it was trained on; and (b) is a better starting point for finetuning on unseen datasets. We show that ColD Fusion outperforms RoBERTa and even previous multitask models. Specifically, when training and testing on $35$ diverse datasets, ColD Fusion-based model outperforms RoBERTa by $2.33$ points on average without any changes to the architecture.\footnote{We release the final model as well as iterations and seeds here: \url{https://huggingface.co/ibm/ColD-Fusion}}
\end{abstract}

\section{Introduction}

%



Over the last few years, pretrained language models are changing the landscape of NLP, where finetuning a pretrained model typically yields state-of-the-art performance on a diverse set of NLP tasks \citep{Chen2022RevisitingPT}. Consequently, improving a pretrained model has the potential to boost every model finetuned on it. However, pretraining is often so computationally expensive that practitioners rarely seek to pretrain new models from scratch.

In contrast, finetuning is usually dramatically cheaper, allowing a given pretrained model to be finetuned many times; e.g., there are thousands of finetuned BERT variants on the Hugging Face Hub\footnote{\url{https://huggingface.co/models?search=bert}}. Motivated by this, we study if and how finetuned models can be ``recycled'' to create a better pretrained model \citep[c.f.,][]{raffel2021blog}. To avoid confusion, henceforth we refer to any starting point for finetuning a \emph{base model} and only the vanilla model as the pretrained model.

\begin{figure}[t]
\includegraphics[width=\columnwidth]{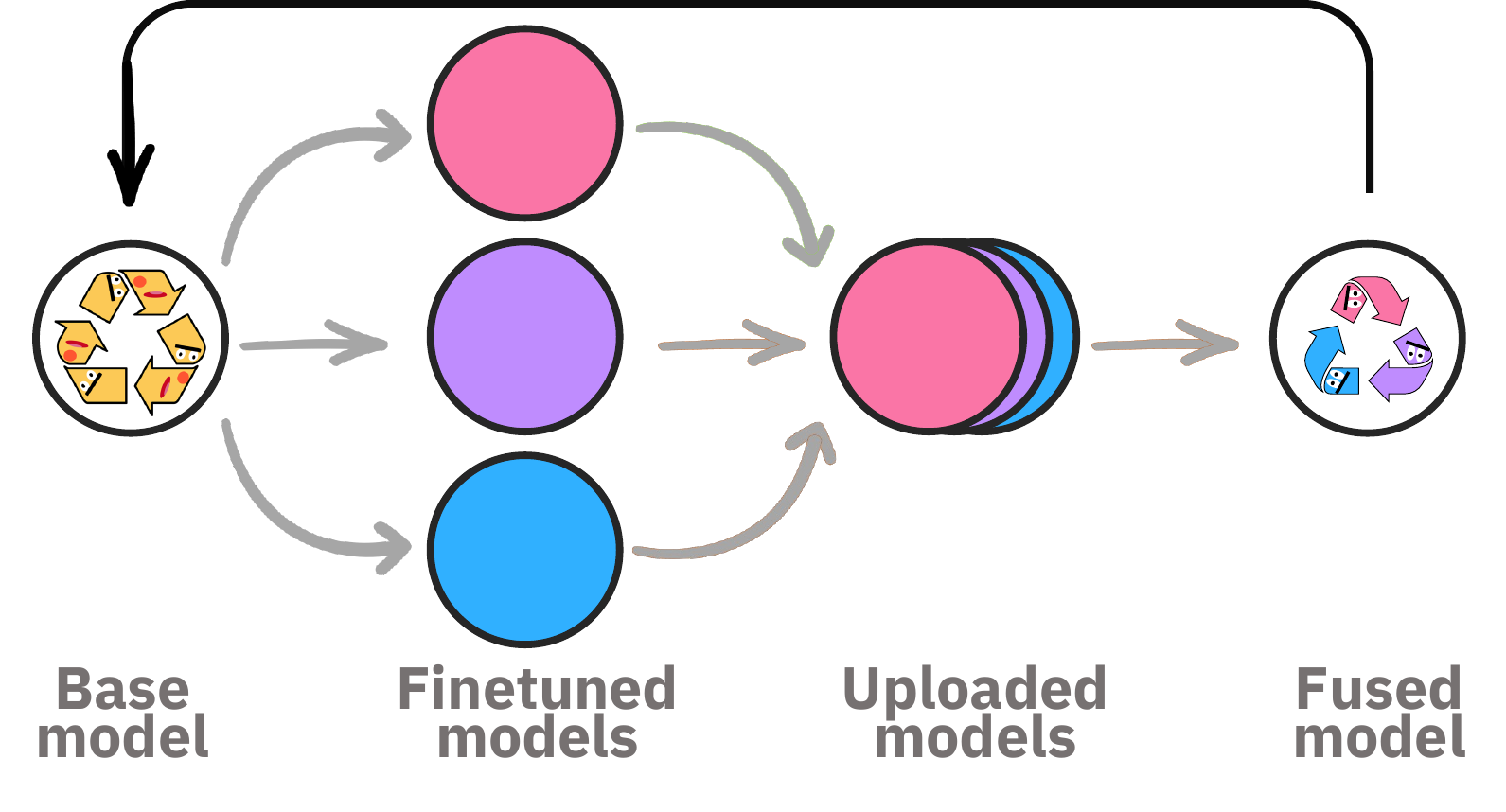}
\caption{Schematic of ColD Fusion. Each iteration starts from a base model. Then, each contributor downloads the model from a centralized ``Repository'' and finetunes it on their dataset. Next, each contributor uploads the finetuned model weights back to the Repository. After that, the Repository fuses all models into a single model by averaging their weights. Finally, the Repository replaces the base model with the fused model and the process repeats. 
}
\label{fig:Schwartz_scheme}
\end{figure}
To recycle models, we take inspiration from multitask learning (\S\ref{sec:background}).
In multitask learning the pretrained model is finetuned over multiple datasets at once, which was shown to create a better base model than the original pretrained model \citep{aribandi2021ext5,aghajanyan-etal-2021-muppet,sanh2021multitaskT0,chung2022scaling}.
Given the availability of many finetuned models, our aim is to obtain the benefits of multitask learning by mixing multiple \textit{models} rather than multiple datasets (c.f. \S\ref{sec:definitions}).

To achieve that, we suggest the following iterative 
approach (\S\ref{sec:methodology}): 
In each iteration, contributors finetune the most up-to-date base model (which is presumably also the most performant) on their task, and share the fine-tuned model with the rest of the community. 
Then, those contributed models are fused together, by simply averaging their parameters \citep{choshen2022fusing}, to create the base model for the next iteration. We call this method Collaborative Descent Fusion, or \emph{ColD Fusion}.
 
ColD Fusion fits the common finetuning paradigm, where each contributor finetunes for their own benefit and does not share their data. However, by merely requiring the finetuned model to be shared, the finetuning step can be recast as a training step for the collective's benefit. In doing so, our method allows reusing compute and data consumed by practitioners and researchers
to the benefit of the entire community. 

Our experimental results indicate that our approach of combining finetuned models not only produces a better base model but also allows this base model to keep evolving. Instead of pretraining or multitasking on a predefined amount of data, we suggest accumulating finetuned models to continuously improve the model. 
Our method is hence limited only by the amount of finetuned models that are shared by the entire community. We discuss limitations in (\S\ref{sec:limitations}). 


We show that ColD Fusion produces a model that performs well on the finetuned tasks, despite never manipulating more than one task at a time neither by constituent models nor their fusing (\S\ref{sec:results}). Moreover, we show that ColD Fusion increases the performance of the base model substantially, outperforming the 
pretrained model by $2.33$ points on average on $35$ datasets. 
Through additional analysis, we further show that similar improvements are achieved regardless of whether the target tasks were seen or unseen during training (\S\ref{sec:unseen}) and that accumulating models trained on additional data provides continuous improvement (\S\ref{sec:analysis}).

\section{Background}\label{sec:background}
We start by motivating the use of further training on diverse data for enhancing the base model abilities (\S\ref{sec:base_model}). Then, we continue with defining our framework's goals (\S\ref{sec:objectives}) and constraints (\S\ref{sec:definitions}).

\subsection{Performance Scaling Laws}\label{sec:base_model}

Extensive evidence suggests that pretraining with more compute \citep{raffel2020exploring} and data \citep{Liu2019RoBERTaAR,hoffmann2022trainingChinchilla,ivgi2022scaling} improves the resulting pretrained model. Moreover, additional supervised data is beneficial even when introduced after the pretraining stage \citep{Phang2018SentenceEO,choshen2022start}. Extending this supervised stage to multitask learning on diverse data sources improves results even further \citep{aribandi2021ext5,aghajanyan-etal-2021-muppet,sanh2021multitaskT0,chung2022scaling}.

We observe that the data used during finetuning is typically not seen during pretraining. 
Therefore, we hypothesize that using a large amount of the data currently used for finetuning may significantly improve the model quality as a base model for future tasks. As training on all the finetuning data directly is infeasible, here we propose an alternative paradigm to test this hypothesis.

\subsection{Goals of Multitask Learning}\label{sec:objectives}
Multitask learning is typically used towards one of two goals: Either to produce a \emph{single model} that performs well on many seen tasks, or to produce a \emph{base model} that 
will perform well on many unseen tasks after adaptation, e.g., via finetuning. 

\paragraph{Single model.} To produce a single  multitask model, one initializes with a base model with $p$ parameters and optimizes the parameters $\theta \in \mathcal{R}^p$ to minimize the loss over a set of datasets $D$. 
This reflects the traditional objective of multitask learning -- to produce a set of weights that performs well on multiple tasks \citep{caruana1997multitask}. 

\paragraph{Base model.} An alternative goal of multitask learning (and the primary goal in our work) is to produce a base model that will attain strong performance after adaptation. Multitask learning does not directly optimize towards this goal, but has been found to do so indirectly \citep{aghajanyan-etal-2021-muppet,liu2022few}.
In this setting, the out-of-the-box performance of the produced model on seen tasks is less important than the performance after finetuning over new tasks, 
i.e., initializing with the found weights $\theta \in \mathcal{R}^p$ and then finetuning on a desired dataset $d'$. 
We do not explicitly state whether $d' \in D$  or  $d'\notin D$, i.e., whether $d$ was used during the multitask training or not. In \S\ref{sec:unseen}, we empirically show that our method works well in both cases.



We note that our formulation sets no restrictions on the datasets group $D$. Thus, a common scenario might be that some datasets do not have the same label space, number of examples, etc. On the other hand, it is also possible that some datasets are complementary samples from a distribution of the same task. In this case, our approach is similar to training this task distributively as in federated learning \citep{yang2019federated} but without communicating every batch. We demonstrate that our approach also works well in this setting in \S\ref{sec:analysis}.

\subsection{Collaborative Constraints}\label{sec:definitions}
In this work, we target the goals of multitask learning discussed above, but focus on a specific setting with additional constraints, which we call \emph{ColD multitask}. The constraints are required to support large-scale collaborative and distributed multitask learning. In our setting, multiple \textit{contributors} have access to datasets that they do not share. A central \textit{Repository} can only perform minimal computation (i.e., does not perform any training). Communication between the contributors and the Repository only occurs when a given contributor completes the finetuning on their data.

\section{Methodology - ColD Fusion}\label{sec:methodology}
Our proposed method (see Fig.~\ref{fig:Schwartz_scheme}), called ColD Fusion, is an iterative process that aims to perform multitask learning in the constrained setting outlined above. Specifically, ColD Fusion involves an iterative process where each individual contributor downloads the current base model from the Repository, finetunes this base model over their dataset, 
communicates the resulting model back to the Repository, and lastly, the Repository fuses \citep{choshen2022fusing} all of the contributors' models into one and sets the new fused model as the new base model for further finetuning.

More formally, the Repository first initializes the shared model parameters $\theta_0$ using a preexisting pretrained model. Then, at each iteration $i \in \{0, 1, 2, \ldots\}$, each contributor $c\in C$ finetunes 
the $\theta_i$ base model over a dataset $d\in D$ 
to produce parameters $\theta_i^c$. 
For the purposes of our study, finetuning is any optimization process that aims to minimize the loss over a 
dataset $d$. Typically, finetuning involves minimizing the loss using a variant of gradient descent. After finetuning, each contributor sends their model's parameters $\theta_i^c$ to the Repository.
Next, the Repository fuses the contributor's models by averaging all of the contributor's model's parameters to produce a new shared model as $\theta_{i+1}= \frac{1}{\abs{C}}\sum_{c}\theta_{i}^c$. Finally, the process repeats for iteration $i + 1$. 

\section{Experimental Setup}\label{sec:exp_setup}
In this section, we detail the datasets, models, baselines, general experiment setup, and specific experiments settings.

\subsection{Datasets}\label{sec:datasets}
In all of our experiments, we define the datasets group $D$ to be a group of 36 English-language datasets, including most GLUE and Super-GLUE datasets, in addition to other NLI, sentiment and topic classification datasets as well as datasets based on Twitter data. A full list of datasets we use is provided in App.~\ref{ap:sec:datasets}. 

At each iteration we test on all the 36 datasets. There are two exceptions: 1) In the main experiment (\S\ref{sec:simulation}) we use the entire dataset group except STSB. STSB, being a regression task incurred technical difficulties to provide a fair comparison to the multitask baseline (see \S\ref{sec:models}). 2). For efficiency reasons, in the very compute demanding experiment of the number of contributors (\S\ref{sec:num_contributors}) we randomly sampled 5 datasets to act as a consistent test set.

\subsection{Models and Baselines}\label{sec:models}
For experiments in the main text, we use RoBERTa-base \citep{Liu2019RoBERTaAR} as our initial model $\theta_0$.
To demonstrate the generality of our approach, we additionally replicate some results on T5 \citep[see App.~\S\ref{sec:t5}]{raffel2020exploring}.

For baseline pre-trained models, we consider RoBERTa-base, RoBERTa-base fused, as well as a RoBERTa-base multitask model. The fused model is trained as in \citet{choshen2022fusing}. The multitask variant trains a dedicated classification head for each dataset. 
In addition, we consider the MUPPET \citep{aghajanyan-etal-2021-muppet} model, a highly optimized multitask model trained on more datasets than we consider. MUPPET is the current state-of-the-art base pretrained model that uses the RoBERTa-base architecture \citep{choshen2022start}.

\subsection{Finetuning Process}\label{sec:finetuning}
Finetuning is used in this paper for two reasons: (a) As a way to infer and evaluate the performance of a base model and (b) as a part of the ColD Fusion scheme. We follow the exact same finetuning procedure in either case. Finetuning hyperparameters and time and memory estimates are provided in App.~\ref{ap:finetune}

\begin{figure*}[t]
\begin{subfigure}{0.48\textwidth}
    \includegraphics[width=\columnwidth]{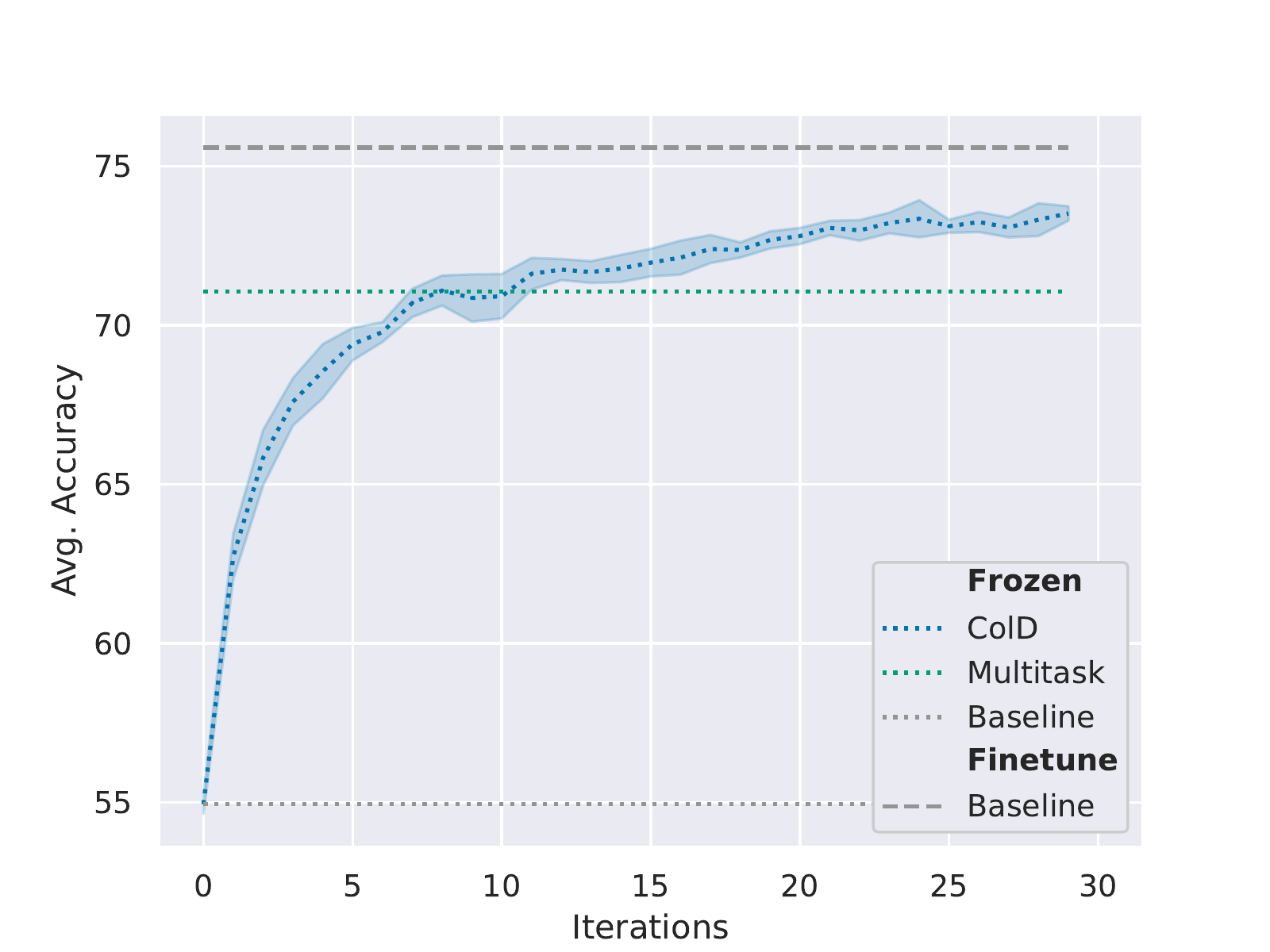}
    \caption{Single Model}
    \label{fig:main_exp_frozen}
\end{subfigure}
\hfill
\begin{subfigure}{0.48\textwidth}
    \includegraphics[width=\columnwidth]{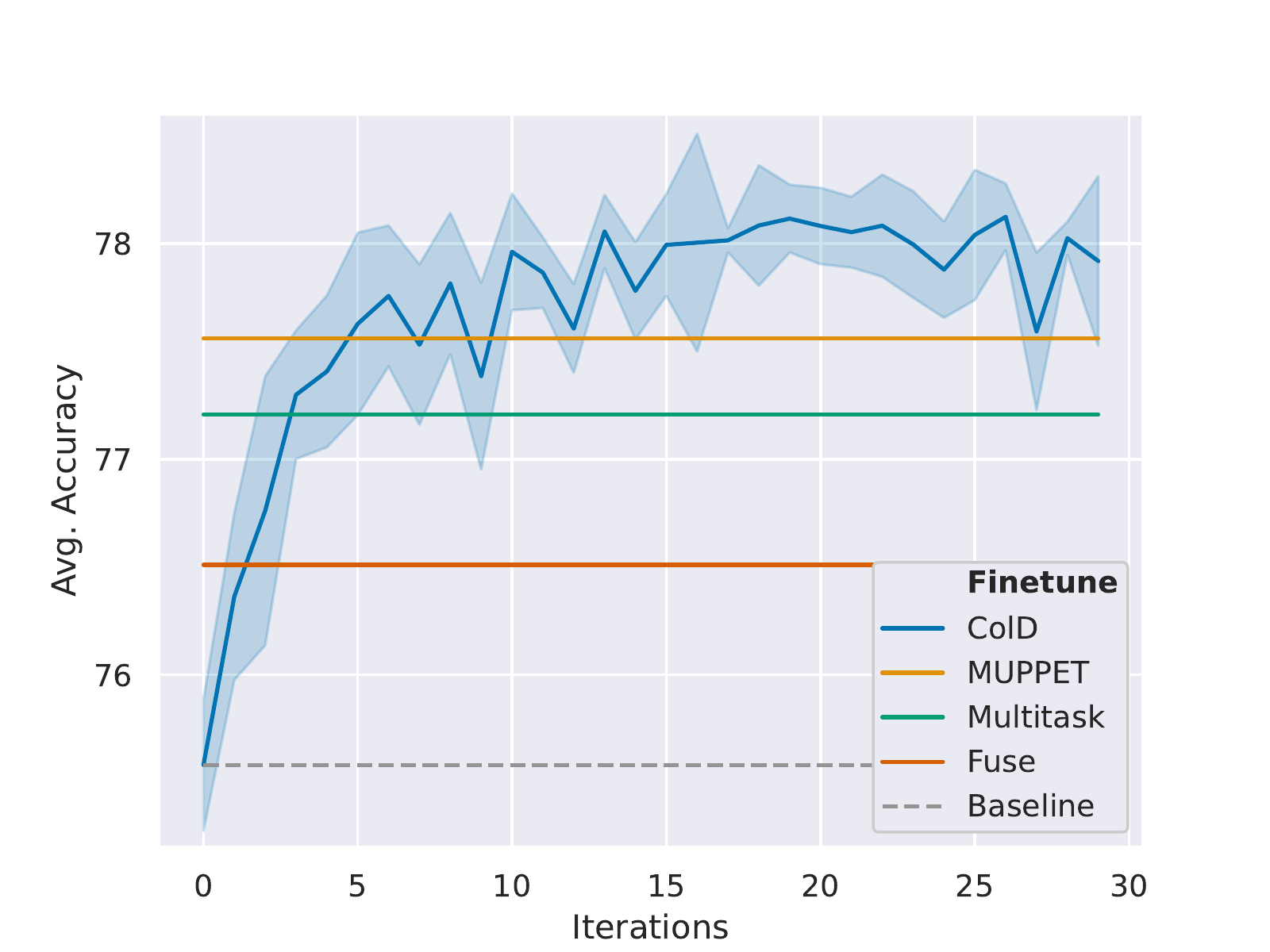}
    \caption{Base Model}
    \label{fig:main_exp}
\end{subfigure}
\caption{ColD Fusion is effective at multitask learning. ColD Fusion brings significant additional benefits as a base model for finetuning (\subref{fig:main_exp}) and improves over finetuning the pretrained model, the fuse baseline, our multitask baseline, and MUPPET  \citep{aghajanyan-etal-2021-muppet}. ColD Fusion also produces better performance on seen tasks as evaluated with linear probing (\subref{fig:main_exp_frozen}), almost reaching finetuned accuracy. Standard deviation across runs is shown via shaded regions.
\label{fig:main_exp_both}}
\end{figure*}

\subsection{ColD Fusion Procedure}
The general course of the experiments is as follows: On each iteration, several datasets are sampled and the latest base model is finetuned separately on each dataset. Then the resulting finetuned models are fused to create the next base model.  This new model is evaluated on the test datasets at each iteration.  
When we mention ColD Fusion without specifying the iteration explicitly, we refer to the model that corresponds to the final iteration. 

The evaluation reflects both multitask goals (\S\ref{sec:objectives}):
(a) To evaluate the single model goal, we train only the classification head \citep[equivalent to Linear Probing;][]{alain2016understanding}, freezing the rest of the layers. We refer to it as ColD-\emph{Frozen}.
(b) For evaluating the base model goal, we take the ColD model and use it as initialization for finetuning. We finetune separately on each dataset and report the results on the corresponding test. We refer to it as ColD.

\section{ColD Multitask Results}\label{sec:results}
In this section, we show ColD Fusion can produce multitask models. We show in \S\ref{sec:simulation} that ColD Fusion fulfills both multitask objectives defined in \S\ref{sec:background}. We verify that improvements replicate on datasets that were not seen during training (\S\ref{sec:unseen}). Then we find that base model improvements are even more apparent in few shot settings (\S\ref{sec:few_shot}). Finally, we consider the importance of the number of contributors hyperparameter (\S\ref{sec:num_contributors}).

\subsection{Collaborative Multitask}\label{sec:simulation} 
We show that ColD Fusion achieves the two multitask objectives (see Fig.~\ref{fig:main_exp_both}). We train and test ColD Fusion for 30 iterations. We simulate 8 contributors by sampling 8 datasets at each iteration and repeat the whole experiment using 5 different random seeds. We consider the importance of the sampling hyperparameter in \S\ref{sec:num_contributors}.

We find that ColD Fusion creates a superior base model (see Fig.~\ref{fig:main_exp}).
The average result after finetuning the ColD Fusion model is superior to the RoBERTa pretrained model by up to 2.33 points on average over the 35 datasets (see App.~\S\ref{ap:improv_datasets} for full results). The result can be deemed significant with a difference of over 20 standard errors of the mean between the original pretrained model and the model produced by ColD Fusion.

In comparison, the standard multitask model and the fused model outperform the original RoBERTa pretrained model by only 1.62  and 0.92 points respectively. We also consider the highly optimized MUPPET model, trained on more datasets and without the ColD multitask restrictions. MUPPET indeed outperforms our standard multitask baseline model, but is outperformed by our ColD Fusion model. 

Another important comparison is the consistency of the improvement. We find (see App.~\ref{ap:improv_datasets}) that the model produced by ColD Fusion is better than the pretrained model on 75\% of the datasets and degrades by only 1.73 points on the worst-case dataset. In contrast, MUPPET hurts as many models as it helps and is worse by 40 points on some datasets.

ColD Fusion also achieves the single model goal: When evaluated with linear probing, the ColD model has high performance on the datasets seen in training (see Fig.~\ref{fig:main_exp_frozen}), higher in fact than those of the standard multitask baseline. Moreover, it is not far from the pretrained model when finetuned on each task separately. This implies that despite learning in a distributed way and fusing by averaging the non-linear weights of the model, the process incorporates the data well.

\subsection{Unseen Datasets}\label{sec:unseen}

\begin{figure}[t]
\includegraphics[width=\columnwidth]{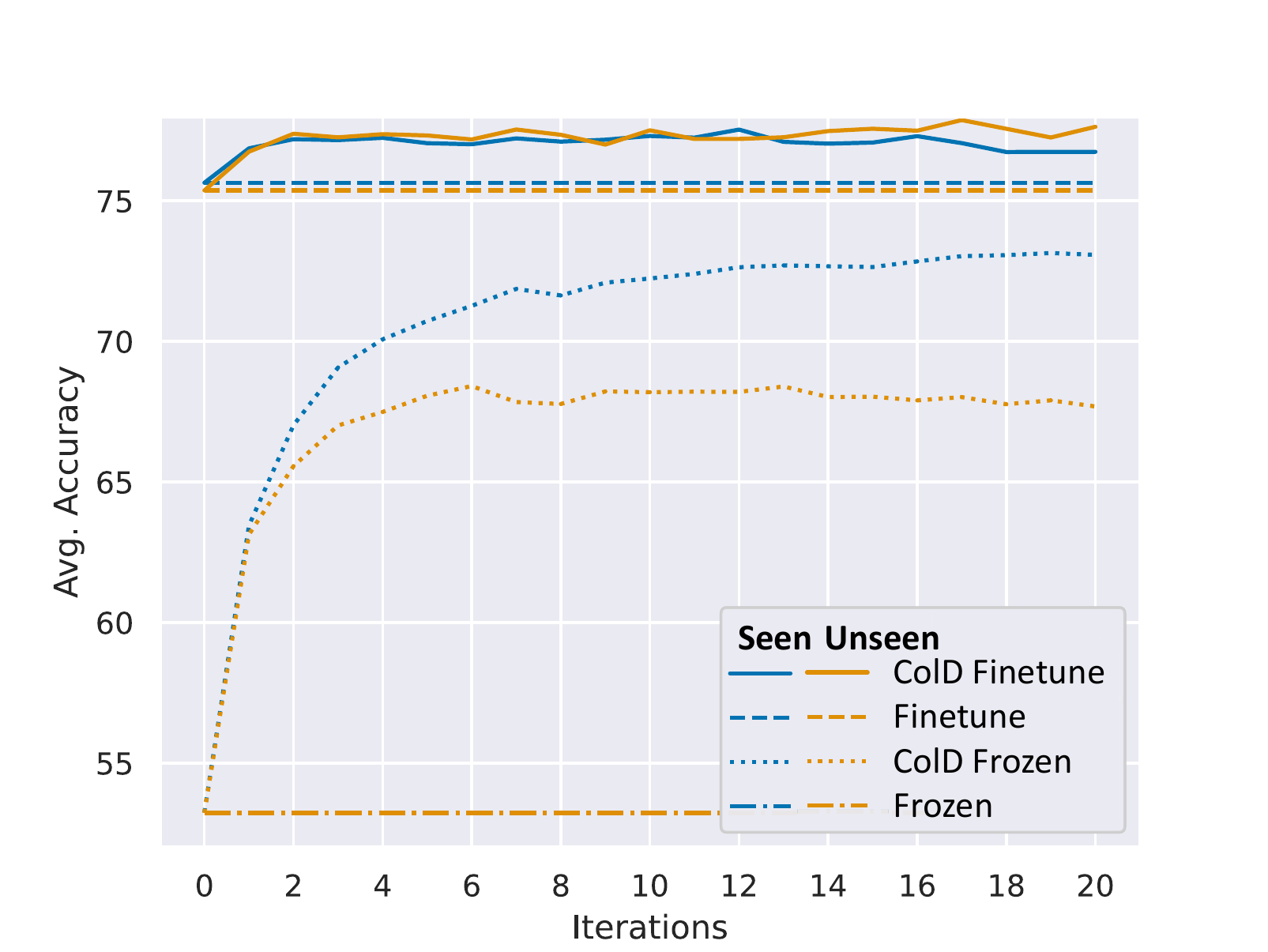}
\caption{Fine-tuned and frozen results for ColD Fusion on datasets that were used for training (``Seen'', in blue) vs. datasets that were not (``Unseen'', in orange). The model produced by ColD Fusion is a good base model for both seen and unseen datasets. While using a frozen model is better for seen datasets, unseen datasets still benefit the ColD Fusion process.}
\label{fig:unseen}
\end{figure}

\begin{figure}[t]
\includegraphics[width=\columnwidth]{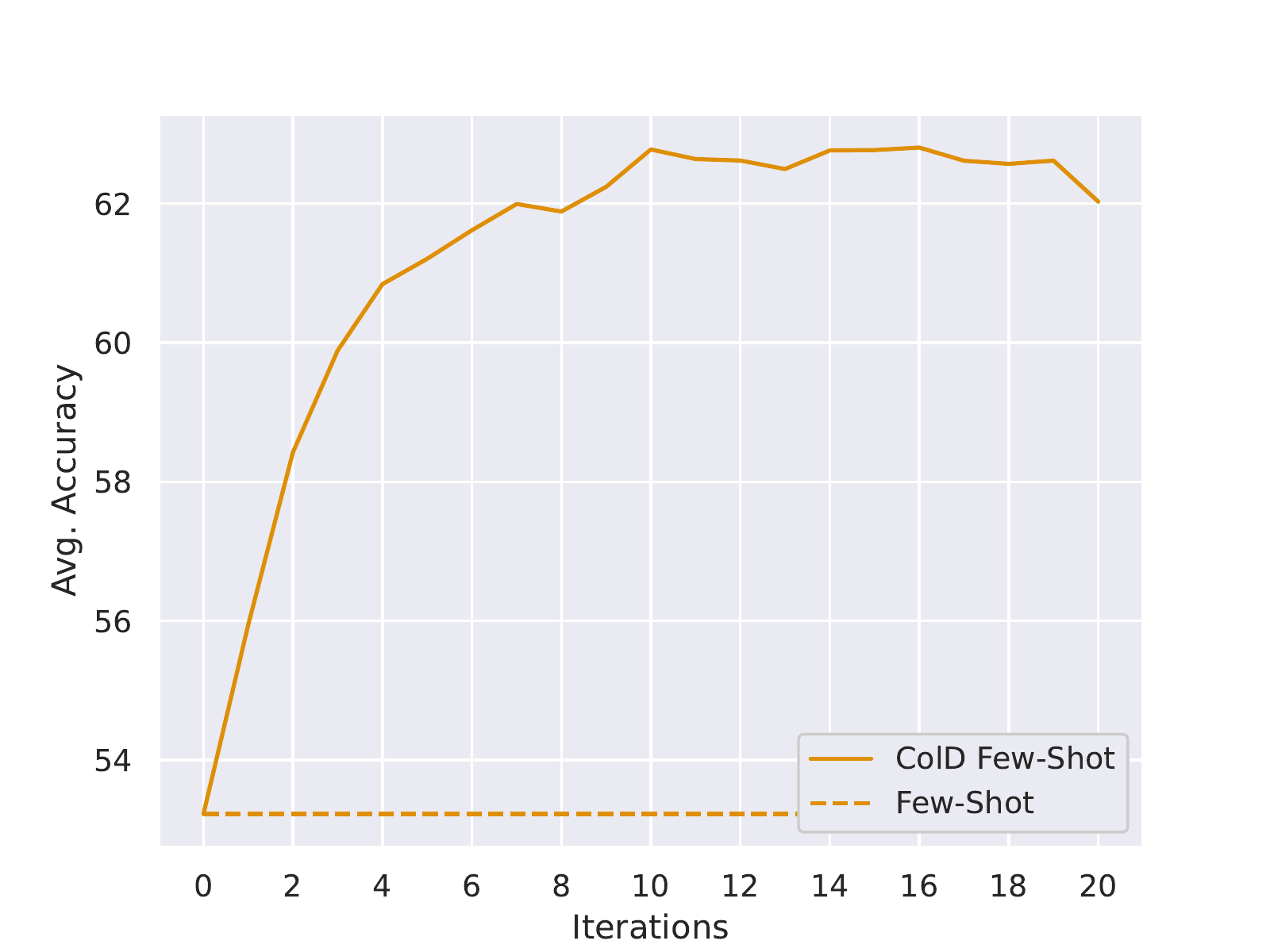}
\caption{ColD Fusion yields improvements in few-shot learning on unseen datasets. Results are from training on 24 full datasets and testing on 12 unseen datasets with 100 labels each, averaged over 3 such folds. The pretrained model performance is
highlighted by a dashed line. ColD Fusion outperforms by as much as 7\%.}
\label{fig:few_shot}
\end{figure}

We have found ColD Fusion to create a strong base model (\S\ref{sec:results}). Next, to meet the requirement of improving results for new datasets, we test the ColD fused model on \emph{unseen} datasets not included in the training (see Fig.~\ref{fig:unseen}). 
We achieve this by performing 3-fold cross-validation. The folds are set arbitrarily such that each fold contains 24 seen datasets (24 contributors) and 12 unseen ones that we keep for evaluation only. This ensures that each dataset has the same weight in the average score of the seen datasets and unseen datasets.

We find that the model performs on unseen datasets just as well as it does on seen ones. 
The strikingly similar performance between seen and unseen tasks (which is similar to in-domain vs.\ out-of-domain) should raise a red flag in most scenarios. However, in the unique scenario of ColD multitasking, it meets our expectations. Both seen and unseen datasets are exposed at some point - either during ColD Fusion iterations (seen datasets only) or during evaluation as a base model (both seen and unseen). Hence, in the seen case, the model trains twice on the same data, first during base model creation and again when evaluating the base model. It is less of a surprise that training twice on the same data doesn't improve results.
The improvement over the original pretrained is likely due to positive transfer across datasets. 

Where finetuning is restricted to only the classification head (ColD-Frozen in Fig.~\ref{fig:unseen}), the model achieves much better performance on the seen datasets than on the unseen datasets. These results are also in line with the fact that the model (apart from the classification head) was never exposed to the unseen datasets, while the entire model's weights were trained on the seen datasets. We further test ColD Fusion's capacity to scale with more data in \S\ref{sec:analysis}. 
We note that the unseen curve consistently increases, which may suggest that the model has acquired general skills. The curve reaches a plateau around the 10th iteration, and then starts to drop a bit. Possibly, due to an overffiting caused by the limited number of seen datasets. 

Note that the scores in Fig.~\ref{fig:unseen} are a bit lower than in the main experiment in Fig.~\ref{fig:main_exp}. This is most likely due to scaling, as here we keep unseen datasets aside and use fewer datasets for training. We show in a controlled experiment in \S\ref{sec:analysis} that using more datasets improves results.


\subsection{Few-shot}\label{sec:few_shot}
In order to assess the benefit of ColD Fusion on few-shot scenarios, we repeat the setting in \S\ref{sec:unseen}, but finetune only on 100 examples from each unseen dataset during evaluation.
Fig.~\ref{fig:few_shot} shows a great increase in performance over the RoBERTa pretrained model, reaching an improvement of $6.73$ points after 20 iterations.
This provides an even stronger case for ColD Fusion in the few-shot setting.

\subsection{Number of Contributors per Iteration}\label{sec:num_contributors}
An important factor in ColD Fusion is the number of contributors in each iteration. Having fewer contributors per iteration implies effectively training on fewer datasets in each iteration; on the other hand, fusing fewer models may give more importance to each.

\begin{figure}[t]
\includegraphics[width=\columnwidth]{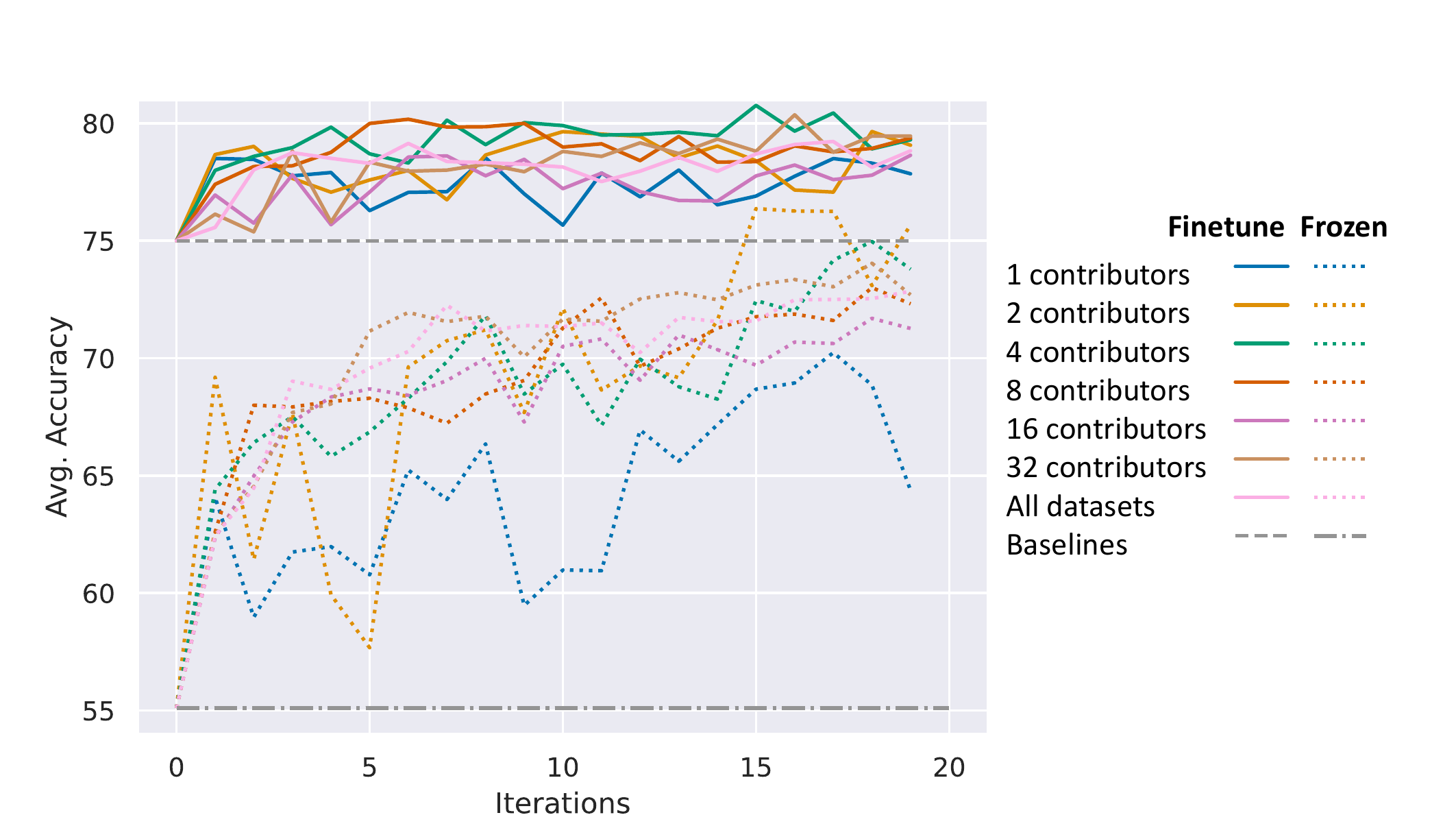}
\caption{Effect of the number of contributors in each iteration. The graph shows the performance (y-axis) per iteration (x-axis) during ColD Fusion. Performance of the model produced by ColD Fusion without additional finetuning is shown in dotted lines, and after finetuning in solid lines. Each color depicts a different number of contributed models in each iteration. The pretrained model performance is highlighted by another dashed line.}
\label{fig:sample_size}
\end{figure}

We observe in Fig.~\ref{fig:sample_size} that starting from two contributors, the performance as a base model is hardly affected by the number of contributors in each iteration. However, adding contributors makes the process more stable. A possible reason is that some of the improvement comes from the iterations themselves and the ability to correct overfitting done in previous steps by some contributors. 

We note that the number of contributors is only insignificant when the data is fixed. In practice, more contributors would improve performance, by adding more data or iterations. We further test the effect of the number of contributors under controlled settings in \S\ref{sec:analysis}.

\section{Single Dataset Analysis}\label{sec:analysis}
We now analyze the interacting effects of the core characteristics of ColD Fusion: additional data across iterations, the amount of training data per iteration, and the number of contributors in each iteration. 

Doing so with multiple datasets would introduce noise. For example, we can not expect additional data coming from different sources (e.g., MNLI or Twitter) to equally affect the performance.
To overcome this, we explore the case where a single dataset is distributed across contributors. Using a single dataset allows us to reduce variability due to differences in the datasets (e.g., distribution, task, etc.), and isolate the parameter we wish to control. ColD Fusion may converge faster with models from a single dataset, but we still expect the general tendencies found to replicate in multiple datasets settings.

We chose MNLI \citep{williams-etal-2018-broad} for its large size (392K examples).



\paragraph{Effect of additional data across iterations (\mbox{Federated Learning}).}
To simulate a never-ending data flow, the experiment runs as follows: at each iteration, 5 contributors sample 5k examples each from MNLI dataset, and another such sample is used for evaluation.

This setting resembles the Federated Learning scenario \citep{yang2019federated}, where multiple contributors collaborate to train a model without having to exchange the actual data.

\begin{figure*}[t]
\begin{subfigure}{0.48\textwidth}
    \includegraphics[width=\columnwidth]{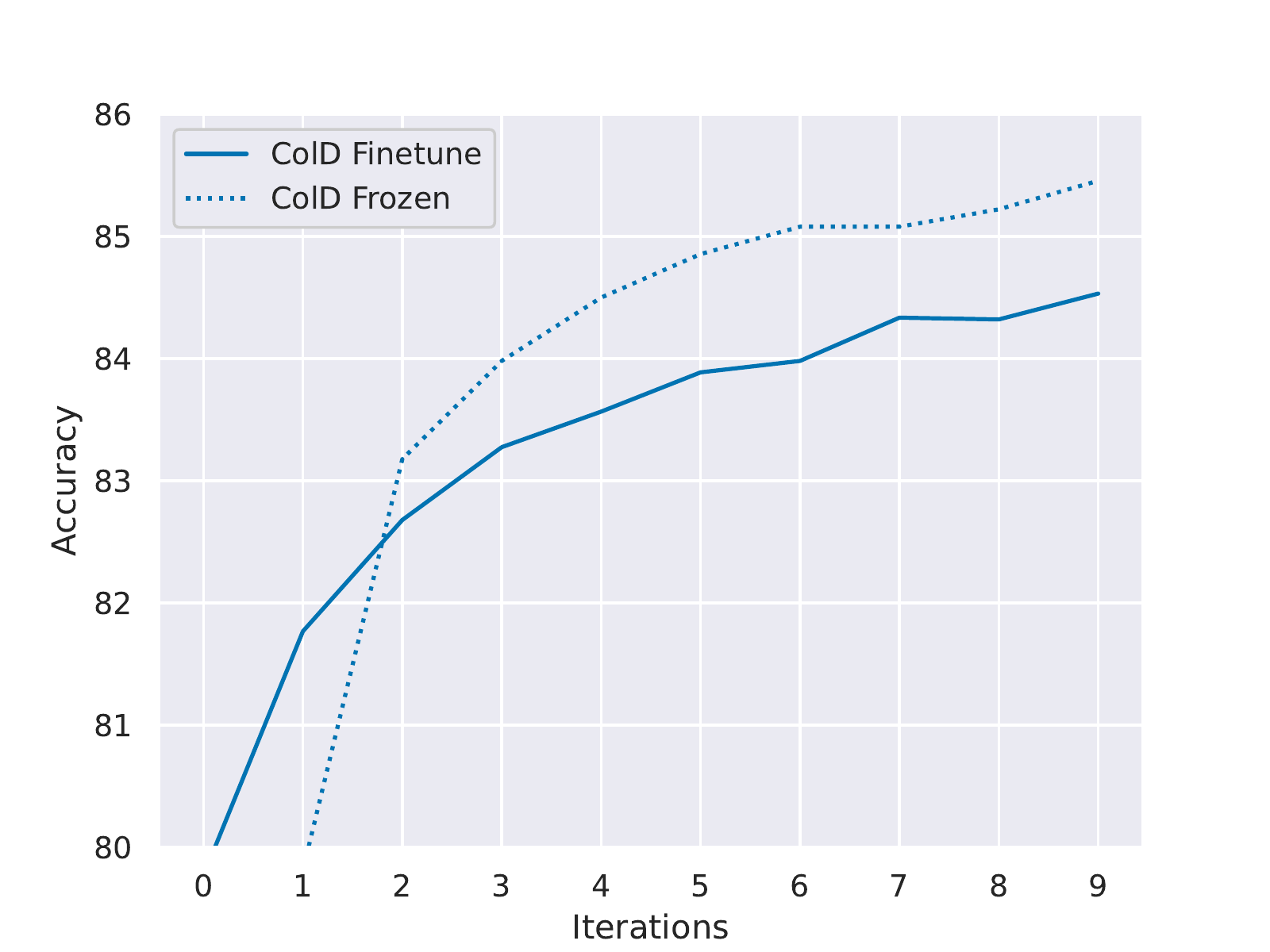}
    \caption{Effect of Additional Data}
    \label{fig:federated_scale}
\end{subfigure}
\hfill
\begin{subfigure}{0.48\textwidth}
    \includegraphics[width=\columnwidth]{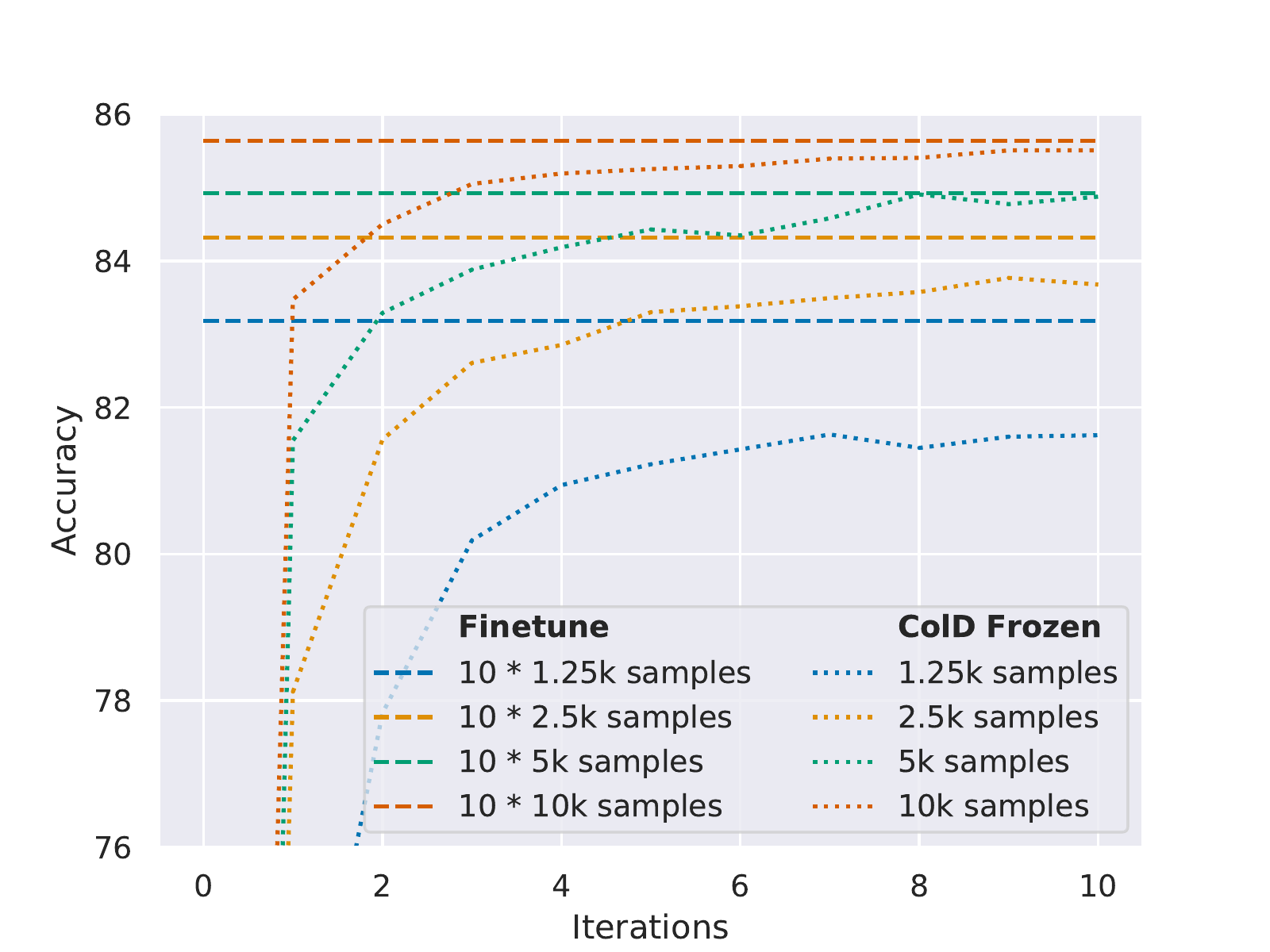}
    \caption{Effect of Dataset Size per Contributor}
    \label{fig:federated_fix_num_models}
\end{subfigure}
\hfill
\begin{subfigure}{0.48\textwidth}
    \includegraphics[width=\columnwidth]{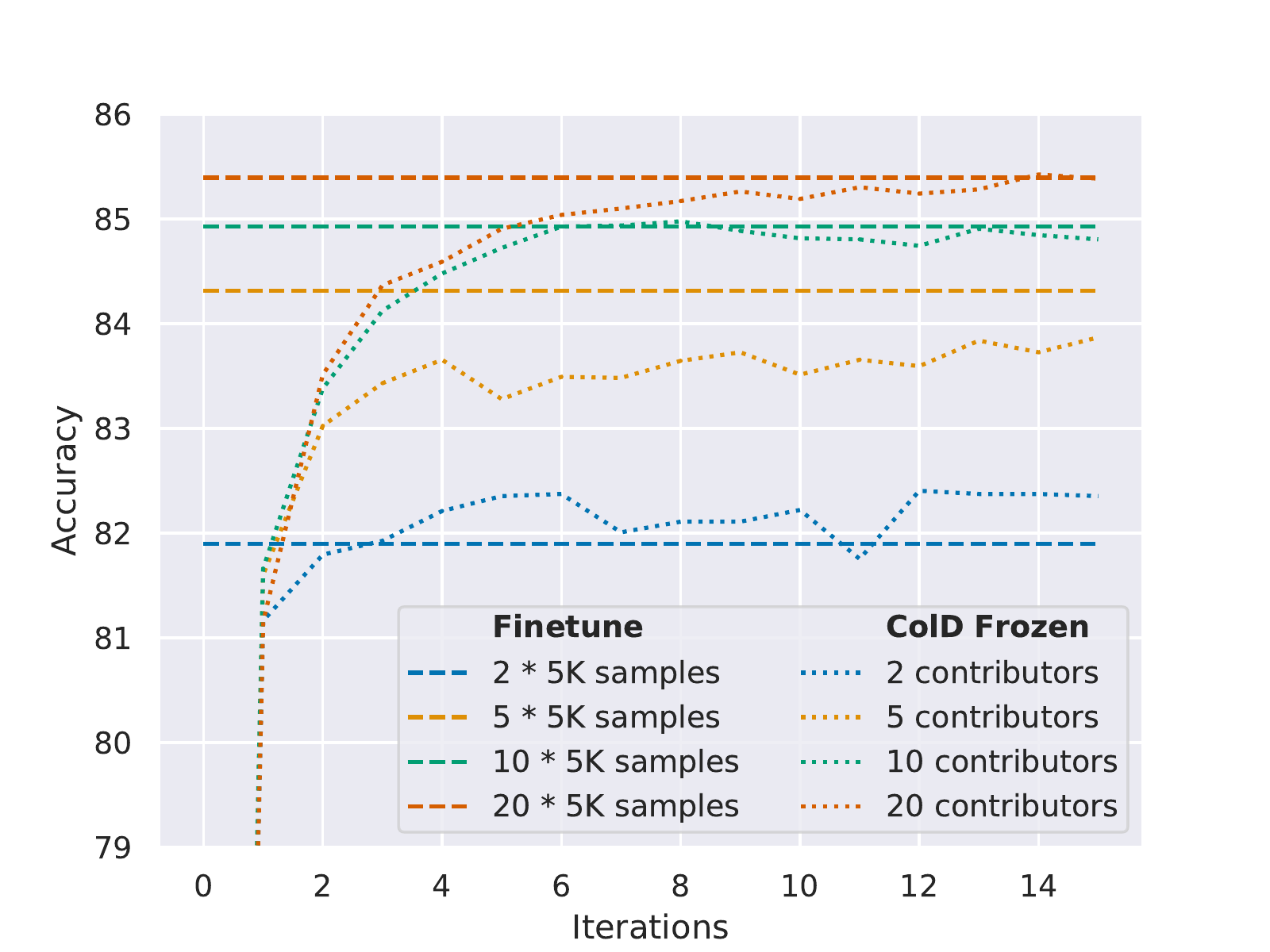}
    \caption{Effect of \#Contributors}
    \label{fig:federated_fix_amount_data_per_model}
\end{subfigure}
\hfill
\begin{subfigure}{0.48\textwidth}
    \includegraphics[width=\columnwidth]{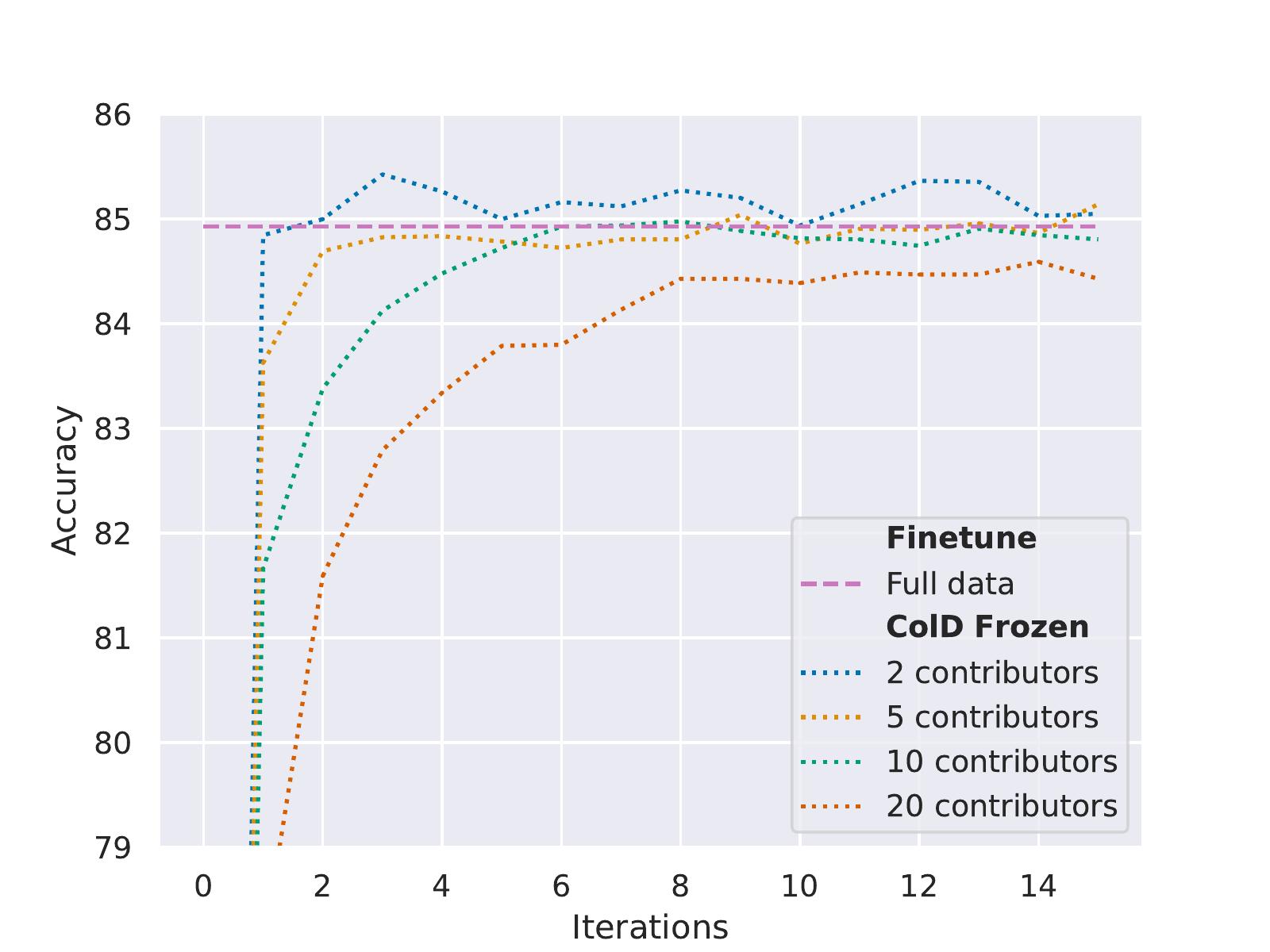}
    \caption{Effect of Distributing}
    \label{fig:federated_fix_total_amount_data}
\end{subfigure}
\caption{ColD Fusion with a single dataset. To test the effect of additional data across iterations (\subref{fig:federated_scale}), we sample 5k new examples from the MNLI dataset for each of the 5 contributors at each iteration. We test the effect of the dataset size (\subref{fig:federated_fix_num_models}), the number of contributors (\subref{fig:federated_fix_amount_data_per_model}), and the distributing of a fixed amount of data (\subref{fig:federated_fix_total_amount_data}). The ColD Frozen (dotted lines) outperforms ColD Finetuned (solid lines), possibly due to overfitting, and both improve with the iterations. Results keep increasing with data size. More contributors or less data per contributor slow the convergence to centralized finetuning (dashed lines).
\label{fig:analysis_both}}
\end{figure*}

As presented in Fig.~\ref{fig:federated_scale}, performance increases throughout the iterations. Thus, we conclude that the ColD Fusion scheme aggregates and utilizes the newly added examples and not only coarse-grained dataset characteristics. 

We show similar trends in the multitask scenario (see App.~\ref{ap:sec:multitask_scale}). Training on more datasets results in a better best model at the cost of more iterations to get to that best model.

Note the superiority of ColD-Frozen over ColD in this experiment. A possible explanation is overfitting. In evaluation, finetuning all the parameters on only part of the data is worse than keeping the fused weights that are trained on several splits. 

\paragraph{Effect of dataset size per contributor.}
In this and the following experiments, we train on all the data in each iteration. The contributors train over disjoint and consistent sub-datasets, i.e., we do not sample examples. We aim to analyze the ability of the model to aggregate knowledge from the constituent models during fusion. 

ColD-Finetuned is evaluated through a stage of finetuning which further learns on the task. To avoid entangling the capabilities learnt during ColD Fusion with those learnt during evaluation, we analyze the ColD-Frozen instead. 
We also note that during evaluation, the classification head is trained on the training data of the first contributor only (which is the only one in the baseline).

We fix the number of contributors to 10 and test how the number of examples each contributor is training on affects results. We experiment with 1.25K, 2.5K, 5K and 10K examples.  We compare these to full finetuning on the union of all the contributors' training data. 
A priori we would have expected large amounts of data in each contributor's model to obstruct the fusing process, as each model changes more. In Fig.~\ref{fig:federated_fix_num_models}, we see the opposite -- the more data each contributor trains on, the closer the fused model is to the full training baseline.  


\paragraph{Effect of the number of contributors.}
In this experiment, each contributor trains over "their own" data, i.e., the same 5K examples in each iteration. We test how the results change with 2, 5, 10 and 20 contributors.
We see in Fig.~\ref{fig:federated_fix_amount_data_per_model} that increasing the number of contributors improves performance. Moreover, the results are not only better at every step, but also keep on improving for longer. This is a positive result in terms of the expected end result, but also means that convergence is slower.

\paragraph{Effect of data distribution between contributors.}
To isolate the effect of the number of contributors and the dataset size of each contributor from that of the overall data size, we fix the overall amount of data to 50K and split it among the contributors evenly. Fig~\ref{fig:federated_fix_total_amount_data} shows distributing mostly affects convergence -- it takes approximately 2 more iterations to converge for double the contributors and half the data seen by each.

We conclude that increasing the overall amount of data improves performance, as may be expected. The distribution of the data between additional contributors has minimal impact on final performance, but may delay convergence. 


\section{Related Work}
Our work strongly relies on model fusion. Model fusion was first introduced as a way to improve pretrained models by \citep{choshen2022fusing}. In parallel, several works 
such as \citep{matena2021merging,Wortsman2022ModelSA} and lately \citep{jin2022dataless,Ram2022PretrainFI} suggested different ways of fusing for other purposes such as improved finetuning. 

Another fusion usage is the stochastic weight averaging, aiming to stabilize the SGD process by averaging multiple points along the SGD trajectory \citep{izmailov2018averaging}. Unlike the previous, this method utilizes only one model and dataset.

Low-communication distributed training was proposed in similar settings to ours. \citet{wortsman2022fi} proposed distributed finetuning and model fusing in order to produce better finetuned models. This suggestion is equivalent to one iteration of ColD Fusion where all models share the same dataset. \citet{li2022branch,together_2022} also share the similarity of distributed training, but during pretraining on unlabeled data.

Understanding why averaging different models improve quality may be related to theoretical works discussing weight and loss spaces. These works state there is a path of minimum loss between models \citep{garipov2018loss} on which the loss along the path is not increasing. \citet{lubana2022mechanistic, benton2021loss, frankle2020linear} claimed that under some constraints, this path is linear, which suggests that fusing the weights could produce a model that retains the capabilities of the fused models. Although different models on the same task may converge to different locations in the loss space without linear connectivity \citep{juneja2022linear}, and although the case of multitask is more complex \citep{mirzadeh2020linear}, we still believe that these works can partially explain why fusing preserves the capabilities gained by the constituent and when it does not that the next iteration fixes it.
\citet{gueta2023knowledge} further suggests the linear connectivity path is merely a line in a whole connected region, future work may tell whether ColD Fusion searches in this region or crosses it to find new ones.

The literature also includes methods for better aligning models during training \citep{javaloy2021rotograd,NEURIPS2020_3fe78a8a,pmlr-v80-chen18a} or after it \citep{ainsworth2022git,jordan2022repair} to aid in fusing. We did not use those as we wanted to reduce the load on the repository and avoid restricting the contributors' finetuning. However, these methods may improve results in ColD Multitask.

We mention that multitask learning does not optimize the base model objective directly (\S\ref{sec:definitions}). Some works aim to do so \citep{bansal2019learning} through meta-learning, finding models that can learn a new task well or efficiently \citep{hospedales2021meta}. REPTILE \citep{Nichol2018reptile} meta learns in a way that resembles ours by iteratively using models trained for several batches.

\section{Conclusion and Discussion}
We proposed a scheme for utilizing abundant finetuned models to enhance a pretrained model. Our approach does not necessitate the sharing of datasets, but rather assumes each contributor solely finetunes on their own dataset. Hence, we believe that applying this scheme as a collaborative pretraining platform is a viable option and that doing so would result in ongoing improvement of base models.

To scale this approach, it would be beneficial if the repository was updated asynchronously, perhaps relying on recent fusing techniques \citep{ilharco2022editing}. 
In the usual finetuning setting, robustness can be improved by tuning batch size and learning rate. In analogy, in ColD Fusion, one can either increase the number of contributors (batch) and/or restrict the effect of each iteration (learning rate) \citep{smith2018bayesian} to improve the process. Following this line, future work may consider regularizing the distance from the pretrained model (learning rate) when a small number of contributors exist (batch) or consider assigning individual weights to each contributor.


There are many hyper parameters 
to optimize which might improve the method substantially. E.g., 
fusing the contributions with a weighted average, improving fusing itself \citep{matena2021merging,ainsworth2022git}, controlling the datasets seen in each iterations \citep[related to;][]{Choshen2021TheGT,hacohen2019power}\cready{Saphra's syntactic interference during training paper?} and backtracking when a harmful update was done to the model.
We hope that future work will shed more light on these issues, to further improve the approach proposed in this work.

\section{Limitations}\label{sec:limitations}
Perhaps the most important limitation regarding ColD Fusion is its deployment. This paper presents a method for multitasking, not a platform. In that sense it solves both multitask learning goals under the constraints resulting from collaboration. However, using ColD Fusion in practice might require much more effort -- It would require a place to host the models, a way to make sure no malicious or erroneous model was sent, and other aspects of a platform to support this training.

This is the first method to tackle collaborative multitasking and we scaled it to 35 datasets. However, future methods may be found more efficient or scale better with the amount of data and computation.

ColD Fusion with many iterations and models might require more computational effort for a given amount of data (\S\ref{sec:analysis}) than regular multitask learning. As a result, while our bottom line performance is encouraging, ColD Fusion might not be the preferred way under every possible scenario. 
Still, some of the costs may be alleviated by future work -- for example the additional iterations when fusing many models, might be reduced by aligning models' weights before fusing \citep{ainsworth2022git}.

While this paper studied the impact of various ColD Fusion parameters, it is unclear how finetuning or even pretraining parameters affect results. However, we do have a reason to believe the method is relatively robust to these refactors through our initial results and the replication on another architecture (App.~\S\ref{sec:t5}).

Another limitation is the assumption that the weights of the model change. Some adaptation methods assume the model is frozen and only its inputs change. In those cases, the model would not be improved by use. Still, even in such cases, multitask learning \citep{wang2023multitask} might be applied on the inputs, or the same model might be used in different ways, where some also adapt parts of it \citep{hulora,jang2023exploring,qin2022exploring, yadav2023resolving}. In those cases, the method might still prove useful, even if it benefits only from some of the contributions.

As mentioned before, another concern is a possible harmful update done by a contributor. Handling it would require monitoring the updates by regularly evaluating the model, or measuring the updates diff to identify noisy models (too large diff / random weights).

\section*{Acknowledgements}

This material is based upon work supported by the National Science Foundation under Grant No. 2145822.

\clearpage
\bibliography{custom}
\bibliographystyle{acl_natbib}

\clearpage
\appendix
\section{Datasets used}\label{ap:sec:datasets}

Most datasets could be downloaded from \href{https://huggingface.co/datasets/}{huggingface datasets}. We explicitly state the download link when relevant.
As we used groups of datasets we report here the full list of datasets they contain.

GLUE: CoLA \citep{warstadt-etal-2019-neural}, SST2 \citep{socher-etal-2013-recursive}, MRPC \citep{dolan-brockett-2005-automatically}, QQP (\href{https://data.quora.com/First-Quora-Dataset-Release-Question-Pairs}{\texttt{data.quora.com/\allowbreak First-\allowbreak Quora-\allowbreak Dataset-\allowbreak Release-Question-Pairs}}), MNLI \citep{williams-etal-2018-broad}, QNLI \citealt{rajpurkar-etal-2016-squad}, RTE \citep{Dagan2005ThePR,BarHaim2006TheSP,Giampiccolo2007TheTP,Bentivogli2009TheSP}, WNLI \citep{Levesque2011TheWS}

SuperGLUE: BoolQ \citep{clark-etal-2019-boolq}, CB \citep{demarneffe:cb}, CoPA \citep{roemmele2011choice}, MULTIRC \citep{khashabi2018looking}, WIC \citep{pilehvar2018wic}, WSC \citep{levesque2012winograd}

MNLI \citep{williams-etal-2018-broad}, QNLI \citealt{rajpurkar-etal-2016-squad}, RTE \citep{Dagan2005ThePR,BarHaim2006TheSP,Giampiccolo2007TheTP,Bentivogli2009TheSP}, WNLI \citep{Levesque2011TheWS}, ESNLI \citep{Camburu2018eSNLINL}, adversarial NLI \citep{nie-etal-2020-adversarial}.

EmoInt \citep{MohammadB17starsem}, Emoji \citep{semeval2018task2}, Irony \citep{van-hee-etal-2018-semeval}, OffenseEval \citep{zampierietal2019}, HatEval \citep{basile-etal-2019-semeval}, Sentiment Analysis \citep{rosenthal-etal-2017-semeval}

Poem Sentiment \citep{sheng-uthus-2020-investigating}, IMDB \citep{maas2011imdb}, Rotten Tomatoes \citep{pang-lee-2005-seeing}, SST 5bins \citep{socher-etal-2013-recursive}, SST2 \citep{socher-etal-2013-recursive}, Amazon reviews \citep{he2016ups} ,Financial Phrasebank \citep{malo2014good}

AG news\citep{zhang2015character}, ISEAR\citep{scherer1994evidence}, Yahoo answers\citep{zhang2015character}, DBpedia\citep{zhang2015character}, 20 newsgroup\citep{zhang2015character}, TREC in both fine-grained and coarse-grained labels \citep[][]{li-roth-2002-learning}

\section{Finetuning details}\label{ap:finetune}
\paragraph{Hyperparameters.} During finetuning, we use the following hyperparameters: learning rate of 5e-5 with linear decay 0.0006 and batch size 256. Early stopping is performed on the development sets if the accuracy improvement after 256K training examples is less than 0.001. All other finetuning hyperparameters are constant across all experiments and follow the original hyperparameters published by \citet{Liu2019RoBERTaAR}. 

\paragraph{Time and Memory.} Most finetuning steps take an hour or less on an A100 GPU. Fusing times are inconsequential. At each iteration all finetuning runs in parallel on all datasets (8 in most cases) and also test finetuning runs in parallel, (36 in most cases). To put it all together, in the main experiment, 30 iterations with 8 contributors, 36 test sets, and 5 seeds, required approximately 4,800 A100 GPU hours and 3.2 TB of memory if all models are to be saved once.

\section{Datasets Accuracy}\label{ap:improv_datasets}
The full results of the main experiment (\S\ref{sec:results}) can be found in Table~\ref{tab:improv_datasets}. It contains accuracy score for each dataset separately.

\begin{figure*}[t]
\begin{subfigure}{0.48\textwidth}
    \includegraphics[width=\columnwidth]{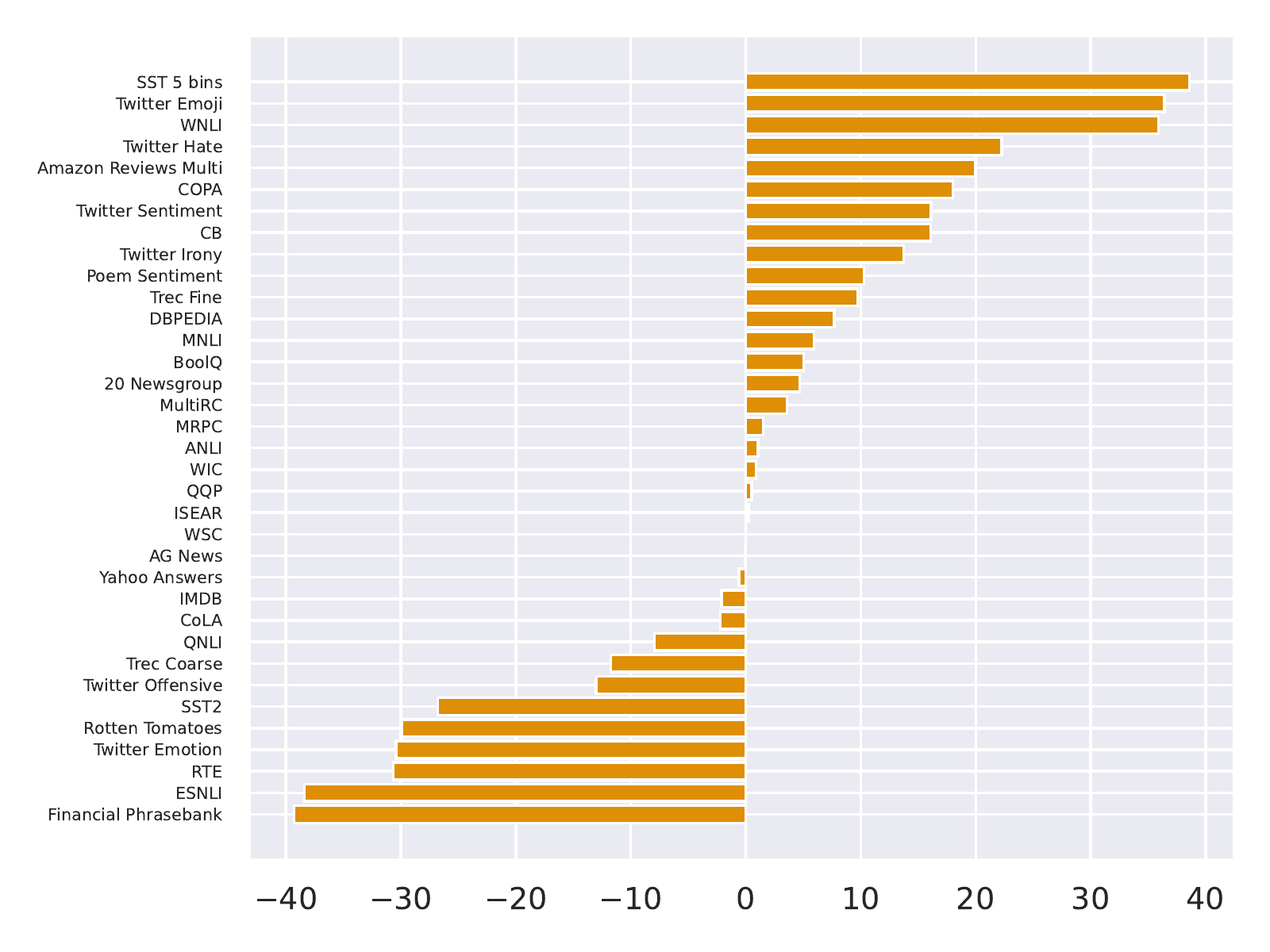}
    \caption{MUPPET Gain}
    \label{fig:muppet_gain}
\end{subfigure}
\hfill
\begin{subfigure}{0.48\textwidth}
    \includegraphics[width=\columnwidth]{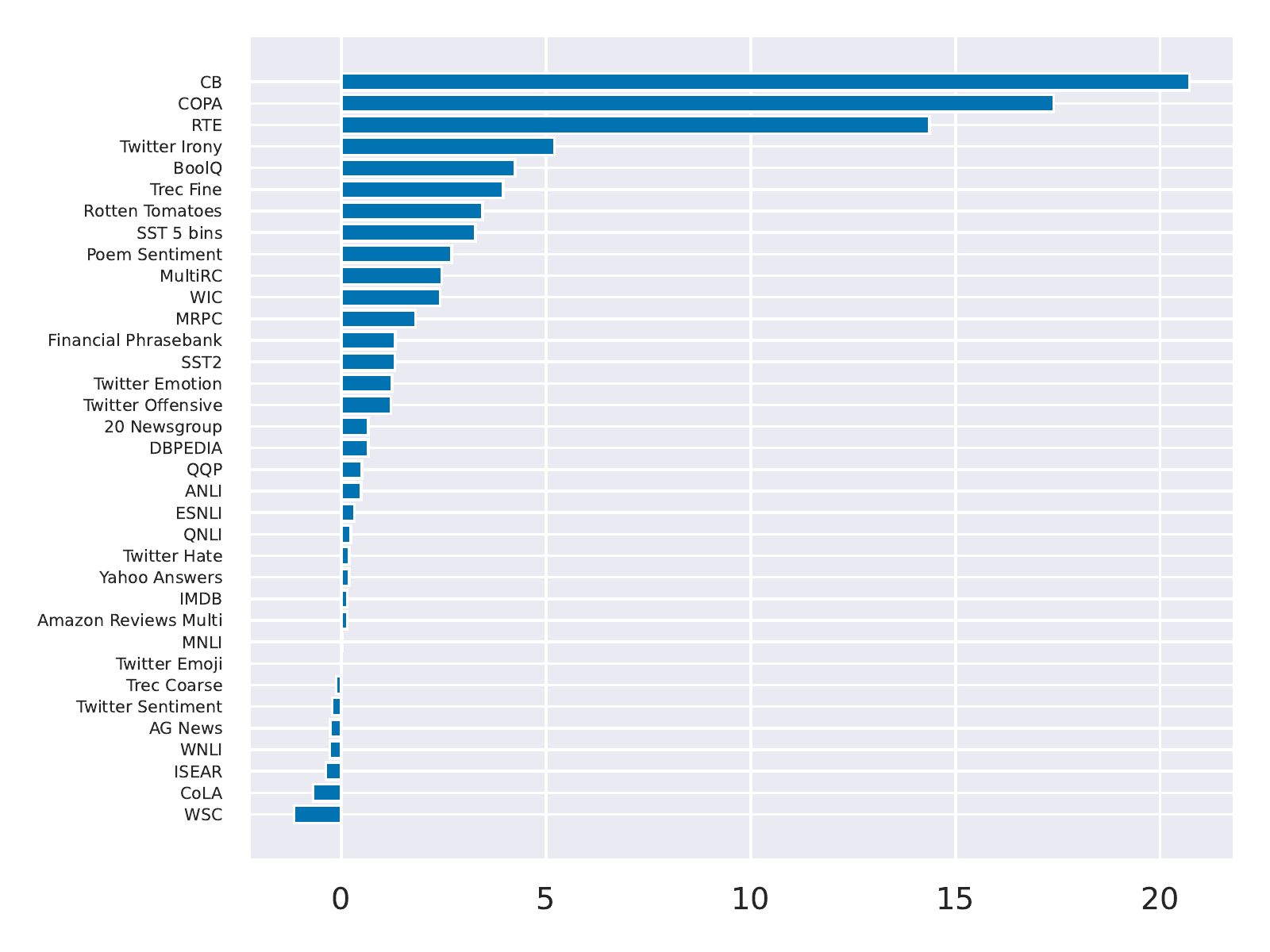}
    \caption{ColD Gain}
    \label{fig:cold_gain}
\end{subfigure}
\caption{
Gains of MUPPET/ColD over finetuning on the pretrained model. ColD is much more consistent, with less datasets that lose from changing from pretrained to ColD and smaller negative effects on them. MUPPET however has larger maximal gain when it does show gains.
\label{fig:barplots}}

\end{figure*}

For ease of comparison we also supply two figures (Fig.\ref{fig:barplots}), comparing MUPPET and COLD multitask models to the pretrained. They show that ColD is much more consistent. It has fewer datasets that lose from changing from pretrained to ColD and smaller negative effects when there are such datasets. MUPPET however also has larger maximal gain when it does show gains, which shines favourably on the average. This makes ColD a better choice for an off-the-shelf model, but gives MUPPET an advantage when one tests a target dataset on several pretrained domains.

\section{T5}
\label{sec:t5}

\begin{figure}[t]
\includegraphics[width=\columnwidth]{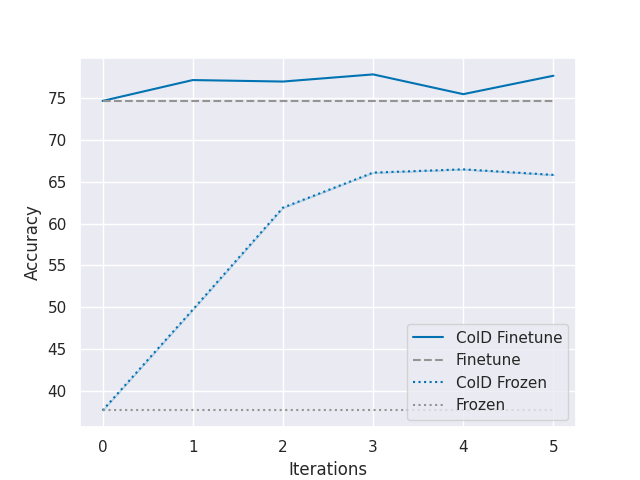}
\caption{ColD Fusion on T5. We replicate the main experiment (\S\ref{sec:results}) on a smaller scale. Like in RoBERTa, both ColD and ColD-Frozen lines keep increasing with the iterations.}
\label{fig:t5}
\end{figure}

We present initial results to confirm our method is not unique to RoBERTa. Specifically, we train T5 \citep{raffel2020exploring} with default hyperparameters, but 256 batch size and 0.0004 learning rate.
We replicate the main experiment (\S\ref{sec:results}) in a smaller scale, running on seed only and 5 iterations only. For ColD-Frozen, we train only the language model head.

Fig.~\ref{fig:t5} shows the main effect reminds. Both ColD and ColD-Frozen keep increasing with the iterations.


\section{Multitask Scale}
\label{ap:sec:multitask_scale}

\begin{figure}[t]
\includegraphics[width=\columnwidth]{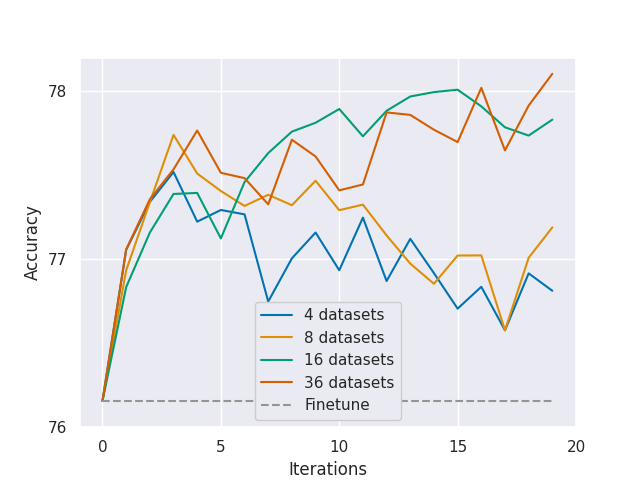}
\caption{Number of datasets effect. The graph follows the performance (y-axis) per iteration (x-axis) during ColD Fusion. Each color depicts a different number of datasets pool from where the datsets were randonly picked at each iteration. The pretrained model performance is highlighted by another dashed line.}
\label{fig:datasets_scale}
\end{figure}

We test the effect of the amount of datasets we use for multitasking on the performance of the resulted model as a base model.
We take a random permutation of all the 36 datasets. We ColD fuse on the first 4 datasets, then the first 8, 16, and finally all the datasets. In fig.~\ref{fig:datasets_scale} we see that the 8 datasets performs worse than the 4 datasets, and that the high regime (16 and 36 datasets) performs much better than the low regime (4 and 8 datasets). These results align with \cite{Aghajanyan2021Muppet} observation that under 15 datasets more datasets decrease the performance, but past some critical point more datasets increase performance.

\section{Fix Number of Examples}
\label{ap:sec:fix_num_exmples}
\begin{figure}[t]
\includegraphics[width=\columnwidth]{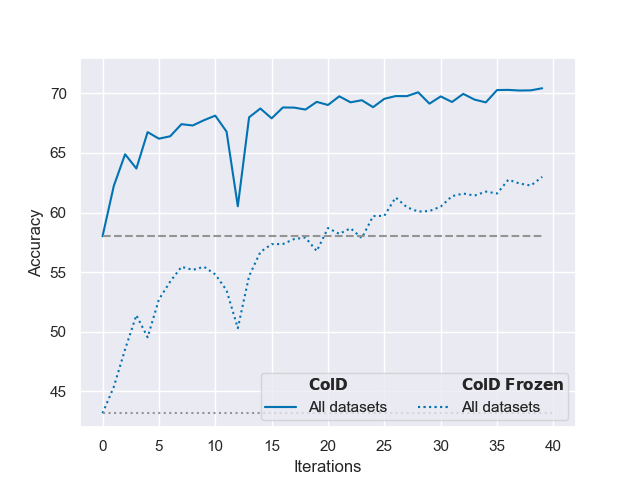}
\caption{Running the main experiment with a fixed number of examples. For each finetune over the 36 datasets, we use 5000 examples - regardless the size of of the dataset. We can see that although the absolute results are degraded related to the regular configuration, the performance is increasing monotonically both for the CoLD and CoLD Freeze. meaning more data yields better performance.}
\label{fig:multitask_fixed_examples}
\end{figure}

We depict the ColD Fusion process with multiple tasks (Fig.~\ref{fig:multitask_fixed_examples}), but only 4K examples per each contributor. This simulates a case where contributors keep streaming new information of different kinds. While this can not fully predict the effect of streaming new tasks, it shows initial positive results in this regard.

\centering
\begin{table*}[htb]
    \centering
    \begin{tabular}{lcccc}
    {Dataset} & {Finetune} & {Multitask}  & {MUPPET} & {ColD-Fusion}  \\
    20 Newsgroup & 85.31 & 85.25 & \textbf{90.00} & 85.97 \\
    AG News & \textbf{89.85} & 89.55 & 89.77 & 89.58 \\
    Amazon Reviews Multi & 66.51 & 66.22 & \textbf{86.50} & 66.65 \\
    ANLI & 51.51 & 51.48 & \textbf{52.59} & 52.00 \\
    BoolQ & 77.14 & 80.27 & \textbf{82.17} & 81.39 \\
    CB & 64.29 & 82.86 & 80.36 & \textbf{85.00} \\
    CoLA & \textbf{83.43} & 82.42 & 81.21 & 82.74 \\
    COPA & 47.00 & 60.00 & \textbf{65.00} & 64.40 \\
    DBPEDIA & 77.49 & 77.69 & \textbf{85.17} & 78.15 \\
    ESNLI & 91.00 & 91.27 & 52.59 & \textbf{91.31} \\
    Financial Phrasebank & 85.40 & 85.26 & 46.10 & \textbf{86.72} \\
    IMDB & 93.86 & 93.82 & 91.74 & \textbf{94.01} \\
    ISEAR & 72.78 & 71.94 & \textbf{73.01} & 72.40 \\
    MNLI & 87.11 & 87.26 & \textbf{93.04} & 87.14 \\
    MRPC & 87.45 & 86.96 & 88.97 & \textbf{89.26} \\
    MultiRC & 60.56 & 62.34 & \textbf{64.15} & 63.01 \\
    Poem Sentiment & 83.85 & 88.27 & \textbf{94.14} & 86.54 \\
    QNLI & 92.42 & 92.39 & 84.48 & \textbf{92.66} \\
    QQP & 90.72 & 90.89 & \textbf{91.25} & 91.22 \\
    Rotten Tomatoes & 88.03 & 90.73 & 58.10 & \textbf{91.48} \\
    RTE & 70.11 & 82.17 & 39.44 & \textbf{84.48} \\
    SST2 & 93.85 & 94.27 & 67.06 & \textbf{95.16} \\
    SST 5 bins & 56.24 & 57.56 & \textbf{94.84} & 59.52 \\
    Trec Coarse & 97.32 & \textbf{97.40} & 85.58 & 97.20 \\
    Trec Fine & 87.08 & 88.28 & \textbf{96.80} & 91.04 \\
    Twitter Emoji & 46.35 & 46.02 & \textbf{82.76} & 46.35 \\
    Twitter Emotion & 81.52 & 81.25 & 51.11 & \textbf{82.76} \\
    Twitter Hate & 53.76 & 53.70 & \textbf{76.02} & 53.95 \\
    Twitter Irony & 71.05 & 74.54 & \textbf{84.77} & 76.25 \\
    Twitter Offensive & 84.58 & 85.16 & 71.57 & \textbf{85.79} \\
    Twitter Sentiment & 70.94 & 70.47 & \textbf{87.07} & 70.72 \\
    WIC & 65.71 & 68.06 & 66.61 & \textbf{68.12} \\
    WNLI & 55.21 & 51.55 & \textbf{91.10} & 54.93 \\
    WSC & \textbf{63.46} & 63.27 & \textbf{63.46} & 62.31 \\
    Yahoo Answers & 72.49 & 71.71 & 71.90 & \textbf{72.69} \\
    \end{tabular}
    \caption{Detailed results of the main experiment. Accuracy score of each dataset, for ColD Fusion and for the 3 baselines: Finetune, our Multitask, and MUPPET.
    \label{tab:improv_datasets}}
\end{table*}
\end{document}